\documentclass[letterpaper]{article}
\usepackage{aaai2026}  
\usepackage{times}
\usepackage{helvet}
\usepackage{courier}
\usepackage{graphicx}
\usepackage{amsmath}
\usepackage{amsfonts}
\usepackage{booktabs}
\usepackage{multirow}
\usepackage{natbib}
\usepackage{booktabs}    
\usepackage{placeins}    

\title{Manifold-Aware Point Cloud Completion via Geodesic-Attentive\\ Hierarchical Feature Learning }

\author{
  Jianan Sun$^{1}$,
  Dongzhihan Wang$^{2}$,
  Mingyu Fan$^{1,*}$\\[6pt]
}

\affiliations {
    \textsuperscript{\rm 1}  Donghua University, Shanghai, China\\
    \textsuperscript{\rm 2}Shanghai University, Shanghai, China\\
}

\begin{document}
\maketitle

\begin{abstract}
Point cloud completion seeks to recover geometrically consistent shapes from partial or sparse 3D observations. Although recent methods have achieved reasonable global shape reconstruction, they often rely on Euclidean proximity and overlook the intrinsic nonlinear geometric structure of point clouds, resulting in suboptimal geometric consistency and semantic ambiguity. In this paper, we present a manifold-aware point cloud completion framework that explicitly incorporates nonlinear geometry information throughout the feature learning pipeline. Our approach introduces two key modules: a Geodesic Distance Approximator (GDA), which estimates geodesic distances between points to capture the latent manifold topology, and a Manifold-Aware Feature Extractor (MAFE), which utilizes geodesic-based $k$-NN groupings and a geodesic-relational attention mechanism to guide the hierarchical feature extraction process. By integrating geodesic-aware relational attention, our method promotes semantic coherence and structural fidelity in the reconstructed point clouds. Extensive experiments on benchmark datasets demonstrate that our approach consistently outperforms state-of-the-art methods in reconstruction quality. 
\end{abstract}









\section{Introduction}

\begin{figure}[t]
  \centering
  \includegraphics[width=0.9\linewidth]{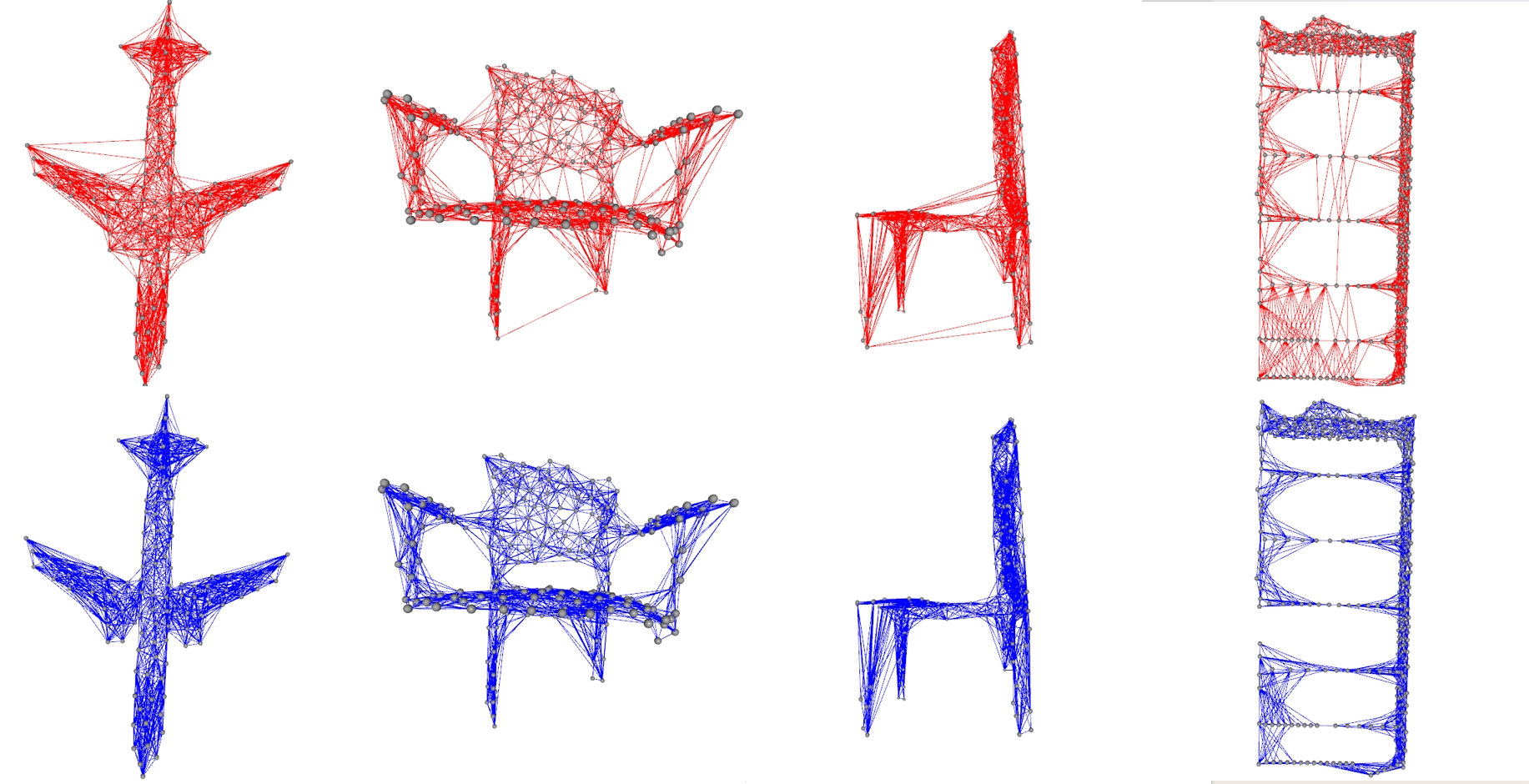}
  \caption{ Visualization of point connectivity using Euclidean versus manifold-based neighbors. Graphs in the top row are constructed using Euclidean neighbors, while the graphs in the bottom row are built by manifold-aware neighbors. The manifold-based graphs better preserve the intrinsic topology and continuity of the underlying shape. }
  \label{fig:euclidean_vs_manifold_resolution}
\end{figure}

In real-world scenarios, 3D scans are often sparse, incomplete, and noisy due to occlusions, sensor limitations, or environmental disturbances, which pose significant challenges to downstream processing and reconstruction~\citep{chang2015shapenet,tchapmi2019topnet,yu2021pointr,xie2020grnet}. Therefore, point cloud completion has became a fundamental task in 3D computer vision, supporting high-level perception and scene understanding in applications such as autonomous driving, robotics, augmented reality, and cultural heritage preservation~\citep{yuan2018pcn,geiger2012kitti,newcombe2011kinectfusion}. 

Recent learning-based approaches have made notable progress in tackling the shape completion problem~\citep{chen2023anchorformer,rong2024cra,ODGNet2024,zhou2025position}. Despite their success, most of these methods treat point clouds as unordered sets in Euclidean space, overlooking the fact that real-world surfaces often lie on low-dimensional manifolds embedded in high-dimensional space~\citep{yu2023adapointr,zhong2025pointcformer}. This oversimplification hinders accurate modeling of intrinsic geometric structures, limiting the capacity to reason about shape continuity beyond local neighborhoods.

Some recent methods incorporate geometry-aware modules for 3D shape modeling. For instance, following the study of DGCNN~\citep{wang2019dgcnn}, AdaPoinTr~\citep{yu2023adapointr} introduces a geometric block that aggregates features from $k$-nearest neighbors ($k$-NN).  PointCFormer~\citep{zhong2025pointcformer} extends local affinity estimation to high-dimensional feature space by introducing residual vectors. However, both approaches still rely on fixed $k$-NN groupings in Euclidean space~\citep{qi2017pointnet++}, making them vulnerable to point clouds with nonlinear geometry. In contrast, PointAttN~\citep{pointatten2024} replaces explicit $k$-NN grouping with global attention weights across all points.

Euclidean $k$-NN graphs usually contain shortcut connections, linking spatially close but geodesically distant points—leading to features from semantically unrelated regions, especially under sparse sampling (as shown in the top row of Figure~\ref{fig:euclidean_vs_manifold_resolution}). To overcome this challenge, we propose a manifold-aware feature extraction framework. The core of the framework is an anchor-based geodesic distance metric, which efficiently approximates true manifold connectivity from partial observations and robustly redefines local neighborhoods (as illustrated in the bottom row of Figure~\ref{fig:euclidean_vs_manifold_resolution}). The geodesic $k$-NN, by comparison, accurately identifies true adjacency points on surfaces, enabling semantic consistency and structural integrity. This geodesic metric serves as the foundation of our framework: it supports geodesic-aware neighborhood grouping, informs attention-based feature aggregation, and facilitates manifold positional embedding. 

In summary, our contributions are threefold:
\begin{itemize}
\item We propose an efficient anchor-based geodesic distance approximation method that captures intrinsic surface topology and enables robust neighborhood construction in sparse and geometrically complex point clouds.
\item We develop a unified manifold-aware feature extraction framework that integrates geodesic guidance into neighborhood grouping, attention-based aggregation, and manifold positional embedding, leading to improved geometry encoding and reasoning.
\item We conduct experiments on standard and challenging benchmarks, demonstrating the effectiveness and generalizability of our method, which consistently outperforms state-of-the-art approaches.
\end{itemize}

\section{Related Work}

\subsection{3D Shape Completion}

Early works such as 3D-EPN~\cite{dai2017shape} adopted volumetric representations but were limited by resolution and scalability. Recently, point-based methods have become the main stream, which are more flexible and memory-efficient. Notable examples include PCN~\cite{yuan2018pcn}, TopNet~\cite{tchapmi2019topnet}, and AtlasNet~\cite{groueix2018atlasnet}, which employed encoder-decoder architectures to directly predict dense point sets. Approaches like FoldingNet~\cite{yang2018foldingnet} and GRNet~\cite{xie2020grnet} further improved the fidelity of output. Despite these advances, most methods are proposed under the assumption of linear locality and overlook the nonlinear distribution of point clouds—leading to limitations in complex or incomplete shapes.


\subsection{Transformer-based Completion Methods}

Transformer-based methods have recently gained much attention due to their strong capacity for global context modeling. PoinTr~\cite{yu2021pointr} regarded the task as a set-to-set translation problem and introduced geometry-aware modules to model local 3D relationships. AdaPoinTr~\cite{yu2023adapointr} introduced adaptive query generation and a denoising strategy to improve robustness and completion accuracy. SnowflakeNet~\cite{Xiang2021SnowflakeNet} introduced a snowflake-like hierarchical decoding process, coupled with a skip-transformer mechanism to propagate geometric patterns across multiple stages. SeedFormer~\cite{Zhou2022SeedFormer} proposed a seed-based hierarchical refinement strategy that recovers fine-grained geometry through a dedicated upsampling Transformer. PointCFormer~\cite{zhong2025pointcformer} employed a relation-weighted attention mechanism and a progressive feature extraction process to jointly capture global context and local geometry. Meanwhile, PointAttN~\cite{pointatten2024} replaced explicit $k$-NN grouping with global attention weights to reduce shortcut edges in Euclidean neighborhoods. Despite their effectiveness, these transformer-based methods typically rely on Euclidean-based neighbor aggregation, which fails to exploit the underlying nonlinear topology for improvements. 

\subsection{Attention Mechanisms}

Some recent approaches have explored geometry-aware attention mechanisms to improve local feature extraction. AdaPoinTr~\cite{yu2023adapointr} incorporates an adaptive geometric block to encode local spatial relationships, while PointCFormer~\cite{zhong2025pointcformer} integrates local affinity estimation through residual vector attention. PointAttN~\cite{pointatten2024} eliminates fixed $k$-NN by applying global attention across all points, thereby alleviating the effect of neighborhood errors. Nonetheless, these attention modules are still rooted in Euclidean or feature-space proximity, which can lead to semantic ambiguity and poor locality preservation—particularly on sparsely sampled surfaces. In contrast, our approach applies manifold-aware attention guided by anchor-based geodesic distances, enabling more faithful modeling of intrinsic nonlinear data structures.

\section{Method}

\begin{figure*}[h]
  \centering
  \includegraphics[width=0.80\textwidth]{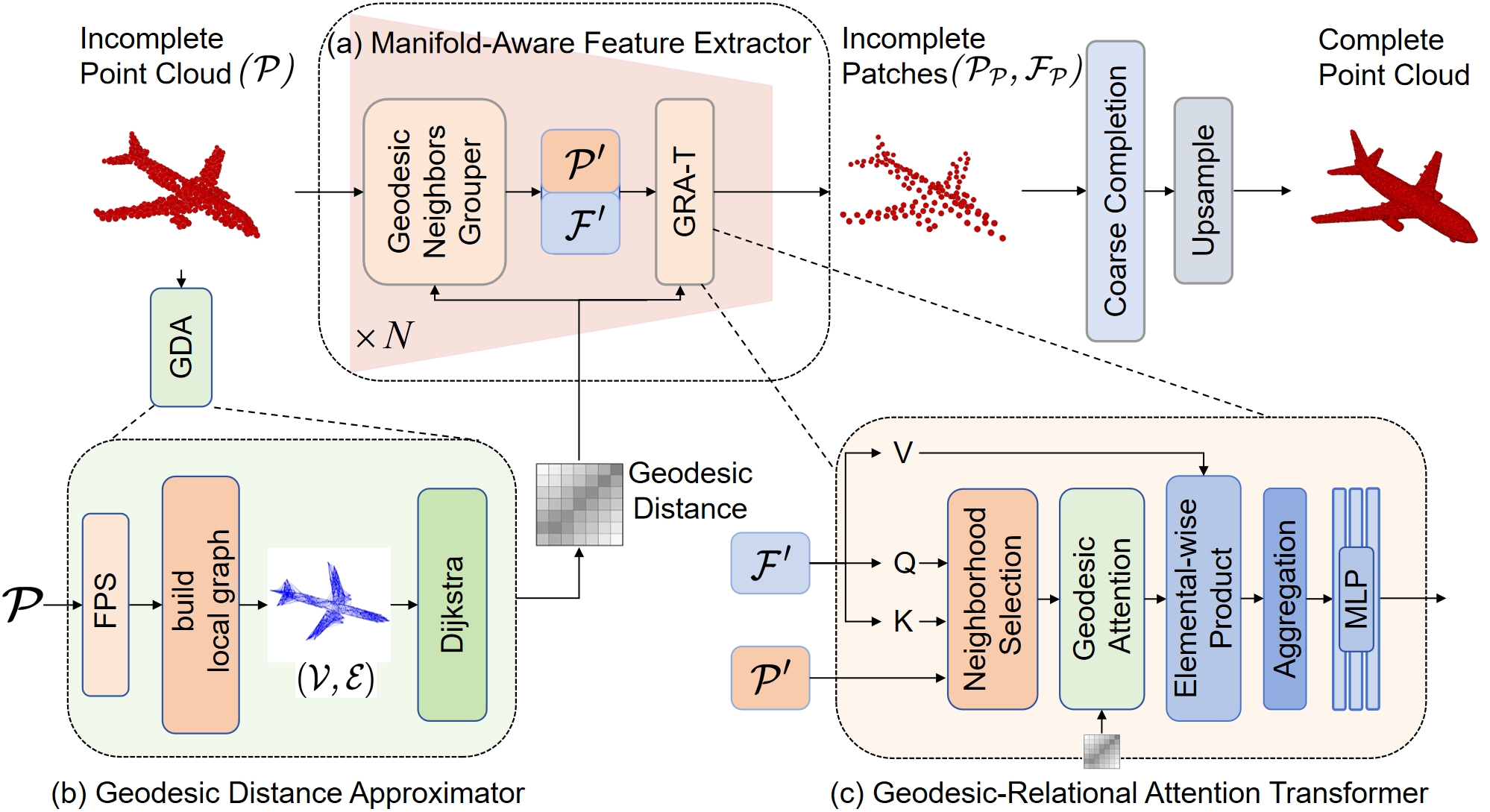}
  \caption{Overview of our proposed framework. Given an incomplete point cloud $\mathcal{P}$, we first construct a local proximity graph and applies Dijkstra’s algorithm to compute geodesic distances in the GDA module. The MAFE module then integrates geodesic information through geodesic-based grouping and our GRA-T module, enabling nonlinear feature aggregation. The resulting features are fed into a coarse completion module, and finally refined into a dense and complete point cloud.}
  \label{fig:framework}
\end{figure*}

\subsection{Overview}

Our approach addresses the point cloud completion task by incorporating manifold awareness into multiple stages of the feature extraction process. In contrast to prior methods that primarily rely on Euclidean proximity while ignoring intrinsic geometric relationships, our pipeline explicitly encodes manifold structures to improve local consistency. As illustrated in Figure~\ref{fig:framework}, the framework comprises three key components: (1) \textbf{Geodesic Distance Approximator (GDA)}: Estimates intrinsic geometric topology by constructing an anchor-based geodesic distance matrix via a sparse graph and shortest-path computation. (2) \textbf{Manifold-Aware Feature Extractor (MAFE)}: Utilizes geodesic $k$-NN grouping to define local neighborhoods and uses the proposed Geodesic-Relational Attention Transformer (GRA-T) to enhance feature aggregation, while a Manifold Positional Embedding (MPE) module injects explicit manifold-aware positional cues. (3) Coarse Completion and Upsample Model: The extracted features are first used for a coarse completion of the global geometric structure, and then refined into a dense and complete point cloud through an upsampling module. By jointly integrating geodesic neighborhood grouping, relational attention, and global manifold position encoding, our method is able to exploit the nonlinear structures of point clouds.

\subsection{Geodesic Distance Approximator (GDA)}

Real-world point clouds often reside on low-dimensional manifolds, where Euclidean distance fails to reflect the true intrinsic geometry. The ideal distance between two points $x_i$ and $x_j$ on a manifold $\mathcal{M}$ is defined as:
\begin{equation}
d_{\mathcal{M}}(x_i, x_j) = \min_{\gamma \in \Gamma_{ij}} \int_0^1 \| \gamma'(t) \| dt,
\end{equation}
where $\Gamma_{ij}$ is the set of all piecewise-smooth paths connecting $x_i$ and $x_j$, and $\gamma'(t)$ denotes the velocity vector along path $\gamma$ at time $t$.
Since exact computation of geodesic distances is intractable for discrete point clouds, we approximate them using a sparse undirected graph $\mathcal{G} = (\mathcal{V}, \mathcal{E})$, where vertices $\mathcal{V}$ correspond to input points and edges $\mathcal{E}$ connect each point to its local Euclidean neighbors. We apply Dijkstra's algorithm \cite{dijkstra1959note} on $\mathcal{G}$ to approximate the geodesic distance matrix $D \in \mathbb{R}^{N \times N}$, thereby capturing the nonlinear manifold structure efficiently.

\subsubsection{Anchor-based Approximation}

To efficiently approximate geodesic distances in large point clouds, we use an anchor-based strategy (in Figure~\ref{fig:framework}(b)). After selecting a sparse anchor set $\mathcal{A}$ via farthest point sampling (FPS), we build a local graph among anchors and precompute geodesic distances $d_{\mathcal{A}}(a_u, a_v)$ between anchor pairs using Dijkstra’s algorithm. This anchor-to-anchor geodesic distance matrix serves as the output of the GDA module and provides the foundation for subsequent geodesic distance estimation between arbitrary points. For any two points $x_i$ and $x_j$, their approximated geodesic distance could be dynamically computed in the subsequent neighborhood grouping and attention computation modules as:
\begin{align}
d_g(x_i, x_j) \approx\; \min_{a_u, a_v \in \mathcal{A}} \big(
    &\, d(x_i, a_u) + d_{\mathcal{A}}(a_u, a_v) \notag\\
    &\,+\, d(x_j, a_v) \big),
    \label{eq:apprx1}
\end{align}
where $d(\cdot, \cdot)$ is Euclidean distance. This anchor-based scheme reduces computational complexity while preserving manifold structure.


\subsection{Manifold-Aware Feature Extractor (MAFE)}
\label{sec:MAFE}

This module incorporates manifold information into a hierarchical feature learning process by combining graph-based neighborhood construction, geodesic-aware attention, and manifold positional embedding, as illustrated in Figure~\ref{fig:framework}(a).

\subsubsection{Geodesic Neighborhood Grouper (GNG)}
This part of our method, GNG, hierarchically downsamples the input point cloud $\mathcal{P}$ using FPS to generate multi-level subsets $\mathcal{P}'$. For each sampled center, we select its $k$ neighbors based on anchor-based geodesic distances rather than Euclidean proximity. The features of these neighbors are aggregated to form local descriptors $\mathcal{F}'$ at each level. This hierarchical grouper realizes multi-scale abstractions and produces features that better capture the intrinsic manifold structure, and serves as the input for subsequent attention modules.

\subsubsection{Geodesic-Relational Attention Transformer (GRA-T)}

\begin{figure}[t]
  \centering
  \includegraphics[width=0.85\linewidth]{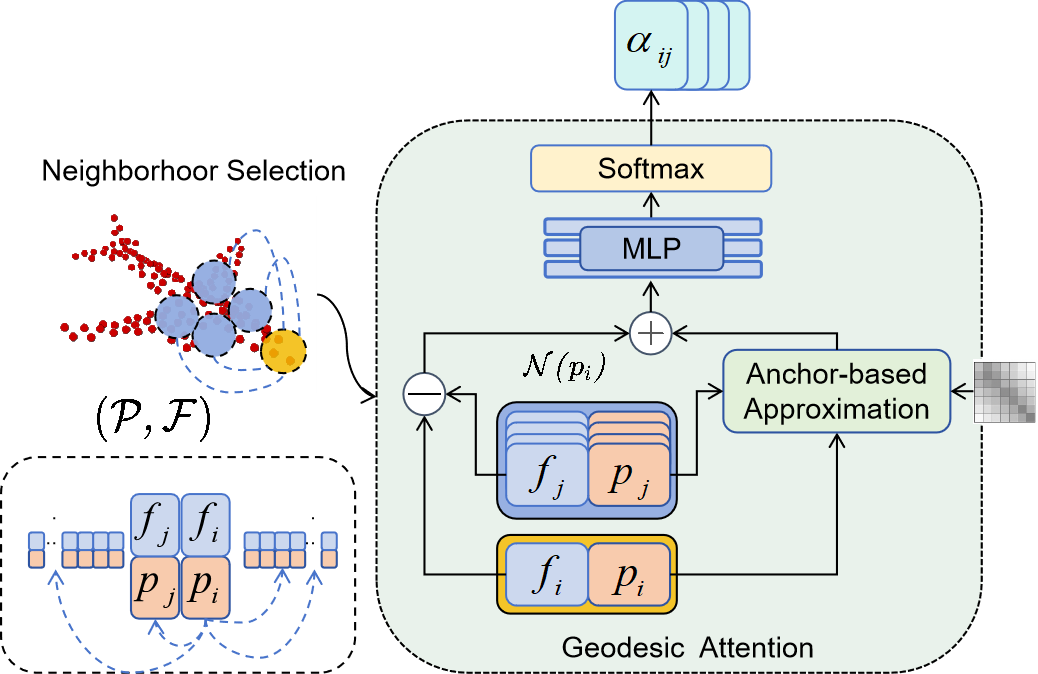}
  \caption{
    Illustration of the geodesic-relational attention mechanism in GRA-T. 
    For each center point, neighborhoods are determined by anchor-based approximation, and both features and coordinates are aggregated using geodesic-aware attention weights.}
  \label{fig:grat}
\end{figure}

To explicitly exploit intrinsic manifold structure, we propose the GRA-T (as illustrated in Figure~\ref{fig:framework}(c)), which integrates geodesic distances into local attention computation, where the geodesic-relational attention mechanism is illustrated in Figure~\ref{fig:grat}. In contrast to conventional attention schemes that rely solely on Euclidean geometry, GRA-T improves local semantic coherence by incorporating geodesic-aware relational cues.
Given a point $p_i$ and its neighborhood $\mathcal{N}(p_i)$ (the set of geodesic-based $k$-NN of $p_i$), we construct geodesic-relational features $r_{ij}$ by combining feature differences and geodesic distances:
\begin{equation}
    r_{ij} = \text{MLP}\left([f_i - f_j, d_g(p_i,p_j)]\right),
\end{equation}
where $f_i, f_j $ denote the features of points $p_i$ and its neighbor $p_j$, respectively; $d_g(p_i,p_j)$ is the precomputed geodesic distance (obtained via the GDA module); and $\text{MLP}(\cdot)$ is a shared multi-layer perceptron. Attention weights $\alpha_{ij}$ are then computed as:
\begin{equation}
    \alpha_{ij} = \frac{\exp(r_{ij})}{\sum\limits_{p_j \in \mathcal{N}(p_i)} \exp(r_{ij})}.
\end{equation}
and used to aggregate the neighbor features into a geodesic-aware refined representation $f_i^{'}$:
\begin{equation} \label{Eq:aggregate}
    f_i^{'} = \sum_{p_j \in \mathcal{N}(p_i)} \alpha_{ij} \odot f_j,
\end{equation}
where $\odot$ denotes element-wise multiplication. 
As demonstrated in Eq.(\ref{Eq:aggregate}), by embedding geodesic relations into the attention computation, GRA-T effectively captures local manifold geometry, enhancing robustness and representational power for feature extraction.

\subsubsection{Manifold Positional Embedding (MPE)}

To further strengthen the manifold-awareness, we introduce MPE that encodes global geometric context into each point’s representation. Specifically, for a point $p_i$, we first identify its closest anchor point $a^* \in \mathcal{A}$, where  $\mathcal{A}$ is the set of predefined anchors. We then retrieve a vector of geodesic distances from $p_i$ to all anchors in $\mathcal{A}$, and concatenate the distances as a positional descriptor to the point feature. Formally, the augmented feature of $p_i$ is expressed as
\begin{equation}
f_i^{\mathrm{aug}} = [f_i,\, d_{g}(p_i, \mathcal{A})],
\end{equation}
where $d_{g}(p_i, \mathcal{A}) \in \mathbb{R}^{|\mathcal{A}|}$ denotes the  geodesic distance vector from $p_i$ to all anchors in in $\mathcal{A}$. This positional embedding provides a compact yet expressive encoding of the point’s global location on the manifold, facilitating better semantic discrimination and geometric consistency.

In summary, the proposed MAFE module, composed of the GRA-T and anchor-based positional embedding, injects intrinsic geometric priors into the feature extraction pipeline. By explicitly modeling both local (geodesic-relational attention) and global (anchor-based position) manifold structures, MAFE significantly improves the model’s representational capacity and robustness.

\subsection{Coarse Completion and Upsample Model}

Following manifold-aware feature extraction, the enhanced point features are used to predict a coarse completed point cloud. We adopt a lightweight coarse-to-fine strategy inspired by SnowflakeNet~\citep{Xiang2021SnowflakeNet}, where deconvolution operations are employed to aggregate geometric cues and generate an initial coarse output. This coarse point cloud is then progressively refined and upsampled by a transformer-based upsampling module~\citep{Zhou2022SeedFormer}, which performs learned geometric refinement and hierarchical feature interpolation. This process enables the reconstruction of dense and semantically consistent point clouds from the initial coarse prediction.

\subsection{Loss Function}
\label{sec:loss}
We adopt the Chamfer Distance (CD)~\citep{cdloss} as the primary loss function to measure the similarity between predicted and ground-truth point clouds, the total loss used to train our manifold-aware completion network is formulated as:
\begin{equation}
    \mathcal{L} = \mathcal{L}_{CD}(P_c, Q) + \sum_{i=1}^{n} \mathcal{L}_{CD}(P_i, Q),
\end{equation}
where $P_c$ denotes the initial coarse output, $P_i$ are the intermediate or refined outputs at different upsampling stages, $Q$ is the ground-truth point cloud, and $n$ is the number of refinement stages.
This multi-stage supervision encourages the network to progressively align all outputs with the target shape, thereby promoting stable training and accurate geometric reconstruction.

\begin{table*}[t]
\centering
\caption{Category-wise CD values ($\times 10^{-3}$) on the PCN dataset. Lower is better.}
\label{tab:pcn_category_results}
\begin{tabular}{l|l|ccccccccc}
\toprule
Methods & Average & Plane & Cabinet & Car & Chair & Lamp & Couch & Table & Watercraft \\
\midrule
FoldingNet \citep{yang2018foldingnet} & 14.31 & 9.49 & 15.80 & 12.61 & 15.55 & 16.41 & 15.97 & 13.65 & 14.99 \\
TopNet \citep{tchapmi2019topnet} & 12.15 & 7.61 & 13.31 & 10.90 & 13.82 & 14.44 & 14.78 & 11.22 & 11.12 \\
AtlasNet \citep{groueix2018atlasnet} & 10.85 & 6.37 & 11.94 & 10.10 & 12.06 & 12.37 & 12.99 & 10.33 & 10.61 \\
PCN \citep{yuan2018pcn} & 9.64 & 5.50 & 22.70 & 10.63 & 8.70 & 11.00 & 11.34 & 11.68 & 8.59 \\
GR-Net \citep{xie2020grnet} & 8.83 & 6.45 & 10.37 & 9.45 & 9.41 & 7.96 & 10.51 & 8.44 & 8.04 \\
SnowFlake \citep{Xiang2021SnowflakeNet} & 7.19 & 4.24 & 9.27 & 8.20 & 7.75 & 5.96 & 9.25 & 6.45 & 6.37 \\
AdaPoinTr \citep{yu2023adapointr} & 6.53 & 3.68 & 8.82 & 7.47 & 6.85 & 5.47 & 8.35 & 5.80 & 5.76 \\
SeedFormer \citep{Zhou2022SeedFormer} & 6.74 & 3.85 & 9.05 & 8.06 & 7.06 & 5.21 & 8.85 & 6.05 & 5.85 \\
PointSea \citep{PointSea} & 6.35 & 3.61 & \textbf{8.54} & 7.33 & 6.58 & 5.21 & \textbf{8.24} & 5.75 & 5.62 \\
CRA-PCN \citep{rong2024cra} & 6.39 & 3.59 & 8.70 & 7.50 & 6.70 & 5.06 & \textbf{8.24} & \textbf{5.72} & 5.64 \\
ODGNet \citep{ODGNet2024} & 6.50 & 3.77 & 8.77 & 7.56 & 6.84 & 5.09 & 8.47 & 5.84 & 5.66 \\
PointAttN \citep{pointatten2024} & 6.86 & 3.87 & 9.00 & 7.63 & 7.43 & 5.90 & 8.68 & 6.32 & 6.09 \\
PointCFormer \citep{zhong2025pointcformer} & 6.41 & 3.53 & 8.73 & \textbf{7.32} & 6.68 & 5.12 & 8.34 & 5.86 & 5.74 \\
\midrule
\textbf{Ours} & \textbf{6.32} & \textbf{3.58} & 8.59 & 7.45 & \textbf{6.56} & \textbf{4.92} & 8.27 & 5.76 & \textbf{5.43} \\
\bottomrule
\end{tabular}

\end{table*}

\section{Experiments}

In this section, we conduct a comprehensive evaluation of the proposed method on several widely-used benchmark datasets. We begin by introducing the datasets and evaluation protocols, followed by quantitative comparisons with state-of-the-art methods. We also provide qualitative visualizations to further demonstrate the effectiveness and generalization capability of our approach.

\subsection{Datasets and Evaluation Metrics}

We evaluate our method on a range of synthetic and real-world datasets, including PCN~\citep{yuan2018pcn}, ShapeNet-55~\citep{yu2021pointr}, ShapeNet-34, ShapeNet-Unseen21, MVP~\citep{MVP}, and KITTI~\citep{geiger2012kitti}.

\textbf{Evaluation Metrics.} We employ the following standard metrics to comprehensively assess completion quality:

\begin{itemize}
    \item \textbf{Chamfer Distance (CD):} Measures the average closest-point distance between the predicted and ground-truth point sets. We report both \( L_1 \) and \( L_2 \) variants as commonly adopted in the literature.
    \item \textbf{F-score:} The harmonic mean of precision and recall, computed with a predefined threshold. It reflects the fidelity and completeness of the reconstructed shape.
\end{itemize}

\subsection{Comparison to the State-of-the-Art}

\begin{table*}[t]
\centering
\caption{Quantitative results on the ShapeNet55/34 \citep{chang2015shapenet} dataset for three difficulty levels. (CD-L2 $\times 10^{-3}$).}
\label{tab:shapenet_results}
\renewcommand{\arraystretch}{0.95}
\setlength{\tabcolsep}{2.5pt}
\small
\begin{tabular}{l|cccc|c|cccc|c|cccc|c}
\toprule
\multirow{2}{*}{Method} &
\multicolumn{4}{c|}{ShapeNet-55} & \multirow{2}{*}{F@1\%} &
\multicolumn{4}{c|}{Seen ShapeNet-34} & \multirow{2}{*}{F@1\%} &
\multicolumn{4}{c|}{Unseen ShapeNet-21} & \multirow{2}{*}{F@1\%} \\
 & CD-S & CD-M & CD-H & CD-Avg &  &
   CD-S & CD-M & CD-H & CD-Avg &  &
   CD-S & CD-M & CD-H & CD-Avg &  \\
\midrule
FoldingNet
    & 2.67 & 2.66 & 4.05 & 3.12 & 0.082
    & 1.86 & 1.81 & 3.38 & 2.35 & 0.139
    & 2.76 & 2.74 & 5.36 & 3.62 & 0.095 \\
PCN
    & 1.94 & 1.96 & 4.08 & 2.66 & 0.133
    & 1.87 & 1.81 & 2.97 & 2.22 & 0.154
    & 3.17 & 3.08 & 5.29 & 3.85 & 0.101 \\
TopNet
    & 2.26 & 2.16 & 4.30 & 2.91 & 0.126
    & 1.77 & 1.61 & 3.54 & 2.31 & 0.171
    & 2.62 & 2.43 & 5.44 & 3.50 & 0.121 \\
GRNet
    & 1.35 & 1.71 & 2.85 & 1.97 & 0.238
    & 1.26 & 1.39 & 2.57 & 1.74 & 0.251
    & 1.85 & 2.25 & 4.87 & 2.99 & 0.216 \\
SnowFlake
    & 0.70 & 1.06 & 1.96 & 1.24 & 0.398
    & 0.60 & 0.86 & 1.50 & 0.99 & 0.422
    & 0.88 & 1.46 & 2.92 & 1.75 & 0.388 \\
AdaPoinTr
    & 0.49 & 0.69 & 1.24 & 0.81 & 0.503
    & 0.48 & 0.63 & 1.07 & 0.73 & 0.469
    & 0.61 & 0.96 & 2.11 & 1.23 & 0.416 \\
SeedFormer
    & 0.50 & 0.77 & 1.49 & 0.92 & 0.472
    & 0.48 & 0.70 & 1.30 & 0.83 & 0.452
    & 0.61 & 1.07 & 2.35 & 1.34 & 0.402 \\
CRA-PCN
    & 0.48 & 0.71 & 1.37 & 0.85 & 0.455
    & 0.45 & 0.65 & 1.18 & 0.76 & 0.451
    & 0.55 & 0.97 & 2.19 & 1.24 & 0.412 \\
PointSea
    & 0.43 & 0.64 & 1.19 & 0.75 & 0.485
    & 0.40 & 0.57 & 1.00 & 0.66 & 0.492
    & 0.50 & 0.88 & 1.92 & 1.10 & \textbf{0.541} \\
AnchorFormer 
    & 0.41 & 0.61 & 1.26 & 0.76 & \textbf{0.558}
    & 0.41 & 0.57 & 1.12 & 0.70 & \textbf{0.564}
    & 0.52 & 0.90 & 2.16 & 1.19 & 0.535 \\
PointCFormer
    & 0.42 & 0.64 & 1.15 & 0.73 & 0.499
    & 0.41 & 0.55 & 1.03 & 0.66 & 0.459
    & 0.53 & 0.88 & 1.97 & 1.12 & 0.420 \\
ODGNet
    & 0.47 & 0.70 & 1.32 & 0.83 & 0.437
    & 0.44 & 0.64 & 1.14 & 0.75 & 0.451
    & 0.59 & 1.01 & 2.26 & 1.29 & 0.415 \\
\midrule
\textbf{Ours}
    & \textbf{0.41} & \textbf{0.57} & \textbf{1.09} & \textbf{0.69} & 0.469
    & \textbf{0.40} & \textbf{0.53} & \textbf{0.99} & \textbf{0.64} & 0.489
    & \textbf{0.45} & \textbf{0.80} & \textbf{1.91} & \textbf{1.05} & 0.415 \\
\bottomrule
\end{tabular}

\end{table*}

\begin{table}[t]
\centering
\caption{Results on the MVP validation set. Both inputs and outputs contain 2048 points. We report CD-$\ell_2$ (multiplied by 10,000) and F-Score@1\% for each method.}
\label{tab:mvp_results}
\begin{tabular}{lcccc}
\toprule
Model        & \#Points & CD-$\ell_2$ & F-Score@1\% \\
\midrule
PCN          & 2048     & 9.77        & 0.321      \\
TopNet       & 2048     & 10.11       & 0.308      \\
PoinTr       & 2048     & 6.15        & 0.456      \\
AdaPoinTr    & 2048     & 4.71 & 0.545 \\
CRA-PCN      & 2048     & 5.33 & 0.529 \\
SymmCompletion    & 2048     & 4.89 & 0.534 \\
\midrule
Ours    & 2048     & \textbf{4.58} & \textbf{0.573} \\
\bottomrule
\end{tabular}

\end{table}

\begin{table}[t]
\centering
\caption{Quantitative comparison on the KITTI dataset in terms of FD and MMD.}
\label{tab:kitti_results}
\setlength{\tabcolsep}{2.5pt}
\begin{tabular}{lcccc}
\toprule \small 
Methods & PoinTr & PointAttN & SymmCompletion & Ours \\
\midrule
FD ($\downarrow$) & \textbf{0.0} & 0.67 & 2.54 & 1.35 \\
MMD ($\downarrow$) & 8.21 & 0.50 & 0.93 & \textbf{0.31} \\
\bottomrule
\end{tabular}
\end{table}

\begin{figure*}[t]
  \centering
  \includegraphics[width=0.9\textwidth]{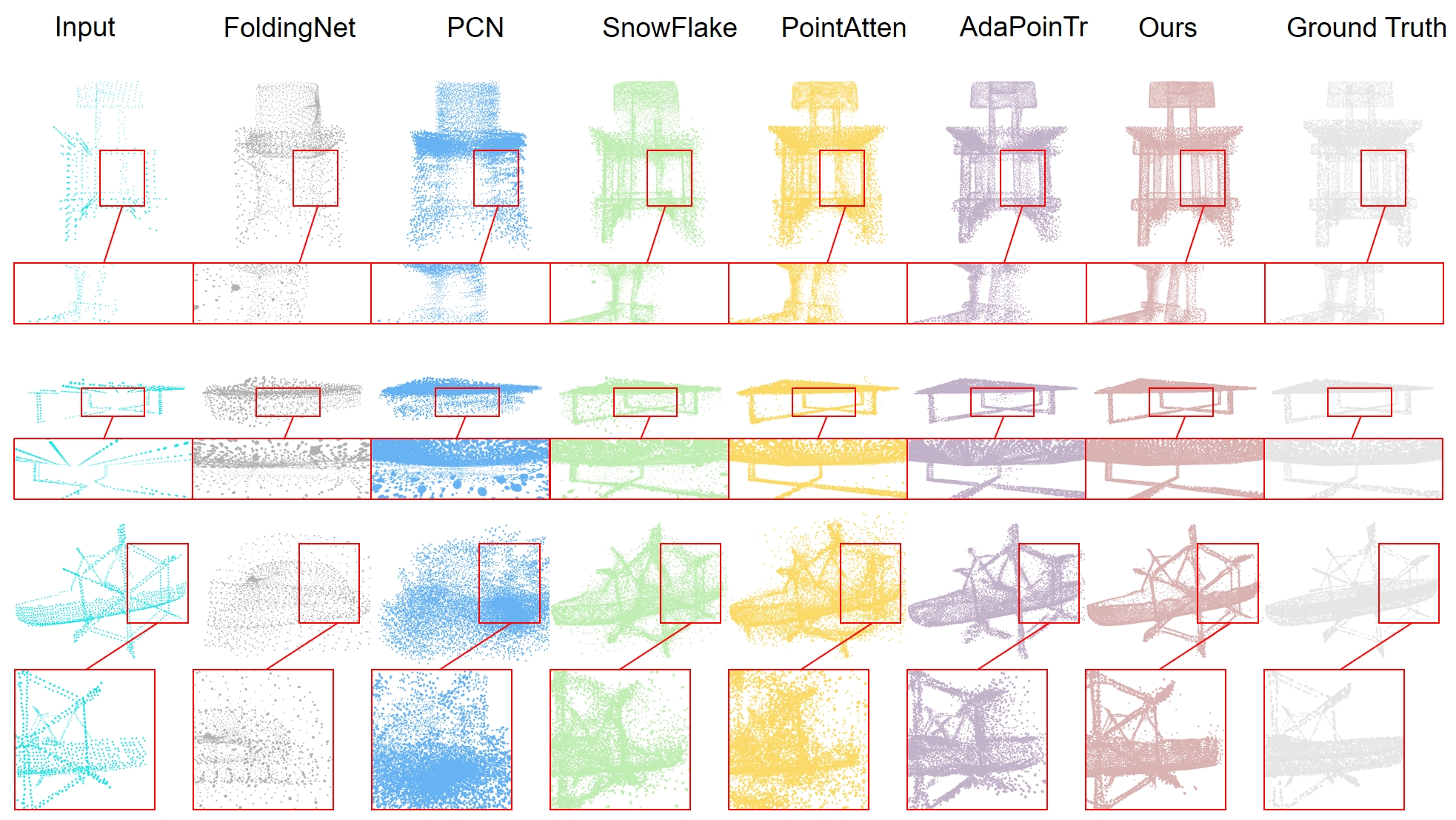}
    \caption{Qualitative results on PCN: our method better restores fine structures and complex topologies compared to prior approaches~\citep{pointatten2024, yu2023adapointr,yuan2018pcn,Xiang2021SnowflakeNet,yang2018foldingnet} .}
  \label{fig:qualitative}
\end{figure*}

Table~\ref{tab:pcn_category_results} reports the category-wise CD values, lower is better, on the PCN dataset. Our method achieves the lowest average CD of 6.32, outperforming all prior state-of-the-art approaches. Notably, our approach obtains the best results in four out of nine categories, including ‘Plane’ (3.58), ‘Chair’ (6.56), ‘Lamp’ (4.92), ‘Watercraft’ (5.43), and achieves competitive performance on other categories, reflecting its strong generalization ability to both simple and complex geometries. Further qualitative results demonstrate the advantages of our approach. For example, as shown in the first example of Figure~\ref{fig:qualitative}, our method successfully reconstructs a chair with multiple thin crossbars, accurately restoring each slat with distinct boundaries. In contrast, previous methods often generate blurred or incomplete connections, failing to preserve the chair’s fine structural details.

Our method achieves state-of-the-art results on ShapeNet-55, consistently outperforming all baseline methods across simple, medium, and hard settings (as shown in Table~\ref{tab:shapenet_results}). Remarkably, despite adopting a more challenging whole-shape completion protocol—where the network must predict the entire object rather than only the missing regions—our model achieves the lowest CD value at every difficulty level~\citep{PointSea}. For example, on the hardest split (CD-H), our method attains a CD of 1.09, a substantial improvement over the best competing methods. As can be seen, our approach exhibits excellent generalization ability, maintaining top performance on both seen and unseen categories, which demonstrates its robustness and applicability to new object classes. Moreover, our method demonstrates strong generalization: it achieves leading results not only on the 34 seen categories but also on the 21 unseen categories, with a CD-Avg of 0.64 and 1.05, respectively. These results confirm the effectiveness of our manifold-aware feature extraction in handling both familiar and novel object classes, and emphasize the broad applicability of our framework to diverse and complex shape completion scenarios.

On the MVP dataset, where both the input and output point clouds consist of 2048 points, our method achieves state-of-the-art results, as summarized in Table~3. Specifically, our approach obtains the lowest CD ($\text{CD-}\ell_2$) of $4.58 \times 10^{-4}$ and the highest F-Score@1\% of 0.573. These results represent a relative improvement of 2.8\% in CD and 2.6\% in F-score compared to the previous best-performing method, AdaPoinTr (CD-$\ell_2$: 4.71, F-Score@1\%: 0.545).

\begin{figure}[t]
  \centering
  \includegraphics[width=0.9\linewidth]{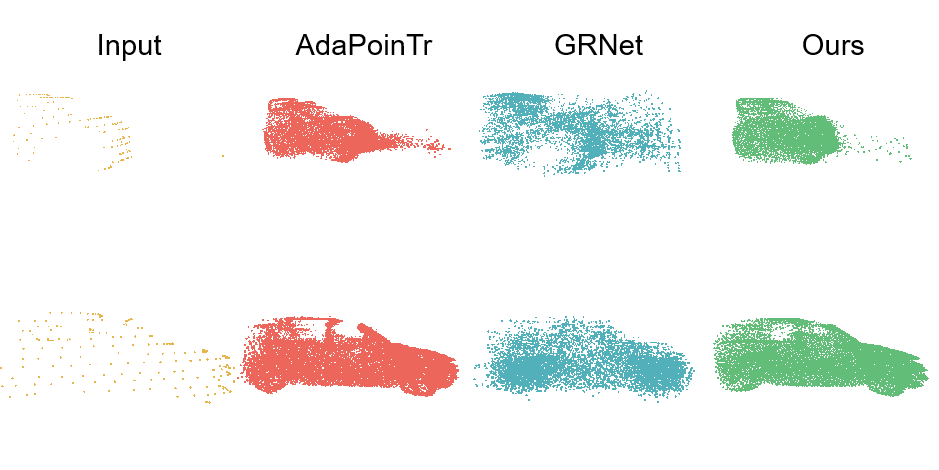}
  \caption{Completion results on the KITTI dataset. Our method recovers plausible shapes under real-world noise and sparsity.}
  \label{fig:kitti_vis}
\end{figure}

Following the evaluation protocol of Zhou \textit{et al.}, we further assess our model—trained solely on the PCN dataset—under realistic conditions characterized by significant noise and sparsity. As illustrated qualitatively in Figure~\ref{fig:kitti_vis} and quantitatively in Table~\ref{tab:kitti_results}, our method delivers competitive performance.

Overall, our method achieves state-of-the-art quantitative results and high-fidelity qualitative reconstructions across all four benchmarks, demonstrating both its effectiveness and versatility.

\section{Ablation Study}

\begin{table}[t]
\centering
\caption{Ablation study of different MAFE module combinations on PCN.}
\label{tab:ablation_combinations}
\renewcommand{\arraystretch}{1}
\setlength{\tabcolsep}{12pt}
\begin{tabular}{cccc}  
\toprule \small 
\textbf{GNG} & \textbf{GRA-T} & \textbf{MPE} & \textbf{CD$\downarrow$} \\
\midrule
-- & -- & -- & 6.74 \\
\checkmark  & -- & -- & 6.42 \\
-- & \checkmark  & -- & 6.39 \\
-- & -- & \checkmark  & 6.64 \\
\midrule
\checkmark  & \checkmark  & -- & 6.41 \\
\checkmark  & -- & \checkmark  & 6.38 \\
-- & \checkmark  & \checkmark  & 6.35 \\
\textbf{\checkmark}  & \textbf{\checkmark}  & \textbf{\checkmark}  & \textbf{6.32} \\
\bottomrule
\end{tabular}

\end{table}

\begin{table}[t]
\centering
\caption{Ablation study: Performance improvement when replacing different backbones' feature extraction modules with MAFE on PCN.}
\label{tab:mafe_backbone}
\begin{tabular}{lcc}
\toprule
Backbone      & Feature Extraction   & CD$\downarrow$  \\
\midrule
SnowflakeNet  & PNet++              & 7.19 \\
SnowflakeNet  & MAFE                & \textbf{6.46} \\
\midrule
SVDFormer     & SVFNet              & 6.54 \\
SVDFormer     & MAFE                & \textbf{6.41} \\
\midrule
SeedFormer    & PNet++              & 6.74 \\
SeedFormer    & MAFE                & \textbf{6.32} \\
\bottomrule
\end{tabular}

\end{table}

\begin{table}[t]
\centering
\caption{Effect of anchor point number on completion performance and inference speed on PCN.}
\label{tab:anchor_ablation}
\begin{tabular}{ccc}
\toprule
Anchor Points & CD$\downarrow$ & Time (ms) \\
\midrule
64    & 6.40          & \textbf{23.11} \\
128   & \textbf{6.32} & 26.58 \\
256   & \textbf{6.32} & 35.02 \\
2048  & --            & 853.12 \\
\bottomrule
\end{tabular}

\end{table}

\subsection{Effect of Different Module Combinations}

To assess the individual and joint contributions of the Geodesic-based Neighborhood Grouper (GNG), Geodesic-Relational Attention Transformer (GRA-T), and Manifold Positional Embedding (MPE), we perform comprehensive ablation experiments, as reported in Table~\ref{tab:ablation_combinations}.
Introducing either GNG or GRA-T alone leads to a substantial reduction in CD, lowering it from 6.74 to 6.42 and 6.39, respectively, relative to the baseline. This indicates that both modules independently enhance local geometric feature extraction. However, combining GNG and GRA-T without MPE results in only a marginal further improvement (CD: 6.41), suggesting partial redundancy in their modeling of local context. In contrast, adding the MPE module yields more pronounced gains when paired with either GNG (CD: 6.38) or GRA-T (CD: 6.35), highlighting the importance of global manifold-aware positional information in resolving local ambiguities. The full configuration, integrating GNG, GRA-T, and MPE, achieves the best performance with a CD of 6.32, demonstrating that these components are complementary and jointly critical for optimal feature representation and shape completion accuracy.

\subsection{Replacing Different Baseline Backbones}

To further demonstrate the generality and adaptability of the proposed MAFE module, we substitute it as the feature extraction component into several representative backbone frameworks, as summarized in Table~\ref{tab:mafe_backbone}.

For both SnowflakeNet and SeedFormer, replacing the original PointNet++ (PNet++)-based feature extractor with MAFE leads to consistent and substantial improvements in reconstruction accuracy, reducing the CD loss from 7.19 to 6.46 in SnowflakeNet and from 6.74 to 6.32 in SeedFormer. For SVDFormer, the improvement is more modest (CD decreases from 6.54 to 6.41). We attribute this to the unique design of SVDFormer, whose feature extractor leverages additional auxiliary information derived from the input point cloud, such as generated multi-view images, to enhance completion.

Overall, the results demonstrate that MAFE offers consistent benefits across diverse backbone architectures.

\subsection{Impact of Anchor Point Number}

We investigate the effect of the anchor set scale on both model accuracy and computational efficiency through controlled ablation studies, with results summarized in Table~\ref{tab:anchor_ablation}. As shown, increasing the number of anchor points from 64 to 128 leads to a noticeable improvement in completion accuracy, with the CD decreasing from 6.40 to 6.32, while maintaining a modest increase in inference time (from 23.11\,ms to 26.58\,ms). Further increasing the anchor count to 256 yields no additional gains in accuracy (CD remains at 6.32) but results in higher inference latency (35.02\,ms), highlighting diminishing returns beyond 128 anchors.

Setting the anchor count equal to the number of input points (2048) effectively removes the abstraction layer, necessitating geodesic distance computations across all points. This configuration is computationally prohibitive, as evidenced by a dramatic increase in inference time to 853.12\,ms and infeasible memory usage, making it unsuitable for practical deployment.

\section{Conclusion}

In this work, we proposed a manifold-aware point cloud completion framework that explicitly integrates geodesic distance estimation and manifold-structured attention into the feature extraction process. By capturing the intrinsic nonlinear geometry of point clouds, our method significantly enhances both geometric fidelity and semantic consistency in reconstructed shapes. Comprehensive evaluations across multiple standard benchmarks confirm that our approach consistently surpasses prior state-of-the-art methods, especially in challenging scenarios with sparse or incomplete observations. We plan to extend our framework to handle dynamic scenes and to explore its integration with multi-modal sensory inputs, aiming to further improve robustness and applicability in real-world environments.


\bibliography{refs/refs}

\end{document}